\documentclass[11pt,letterpaper]{article}
\usepackage{emnlp2016}
\usepackage{times}
\usepackage{latexsym}
\usepackage{setspace}
\usepackage{amsmath}
\usepackage{graphicx}
\usepackage{CJK}
\graphicspath{
    {/}
}

\title{Learning Natural Language Inference using Bidirectional LSTM model and Inner-Attention}

 \author{Yang Liu, Chengjie Sun, Lei Lin\and Xiaolong Wang\\
         Harbin Institute of Technology, Harbin, P.R.China \\ {\tt \{yliu,cjsun,linl,wangxl\}@insun.hit.edu.cn}}

\date{}

\begin{document}
\maketitle

\begin{abstract}
  In this paper, we proposed a sentence encoding-based model for recognizing text entailment. In our approach, the encoding of sentence is a two-stage process. Firstly, average pooling was used over word-level bidirectional LSTM (biLSTM) to generate a first-stage sentence representation. Secondly, attention mechanism was employed to replace average pooling on the same sentence for better representations. Instead of using target sentence to attend words in source sentence, we utilized the sentence's first-stage representation to attend words appeared in itself, which is called "Inner-Attention" in our paper . Experiments conducted on Stanford Natural Language Inference (SNLI) Corpus has proved the effectiveness of "Inner-Attention" mechanism.  With less number of parameters, our model outperformed the existing best sentence encoding-based approach by a large margin.
\end{abstract}

\section{Introduction}
Given a pair of sentences, the goal of recognizing text entailment (RTE) is to determine whether the hypothesis can reasonably be inferred from the premises. There were three types of relation in RTE, Entailment (inferred to be true), Contradiction (inferred to be false) and Neutral (truth unknown).A few examples were given in Table~\ref{RTEexamples}.
  \begin{table}[]
  \small
  \centering
  \label{my-label}
  \begin{tabular}{|l|l|l|}
  \hline
  \textbf{P}    & The boy is running through a grassy area. &   \\ \hline
                & The boy is in his room.                         & C \\ \cline{2-3}
  \textbf{H}    & A boy is running outside.                       & E \\ \cline{2-3}
                & The boy is in a park.                           & N \\ \hline
  \end{tabular}
  \caption{Examples of three types of label in RTE, where P stands for Premises and H stands for Hypothesis}
  \label{RTEexamples}
  \end{table}
  \begin{figure*}[htbp]
  \centering
  \includegraphics[height=8cm, width=16cm]{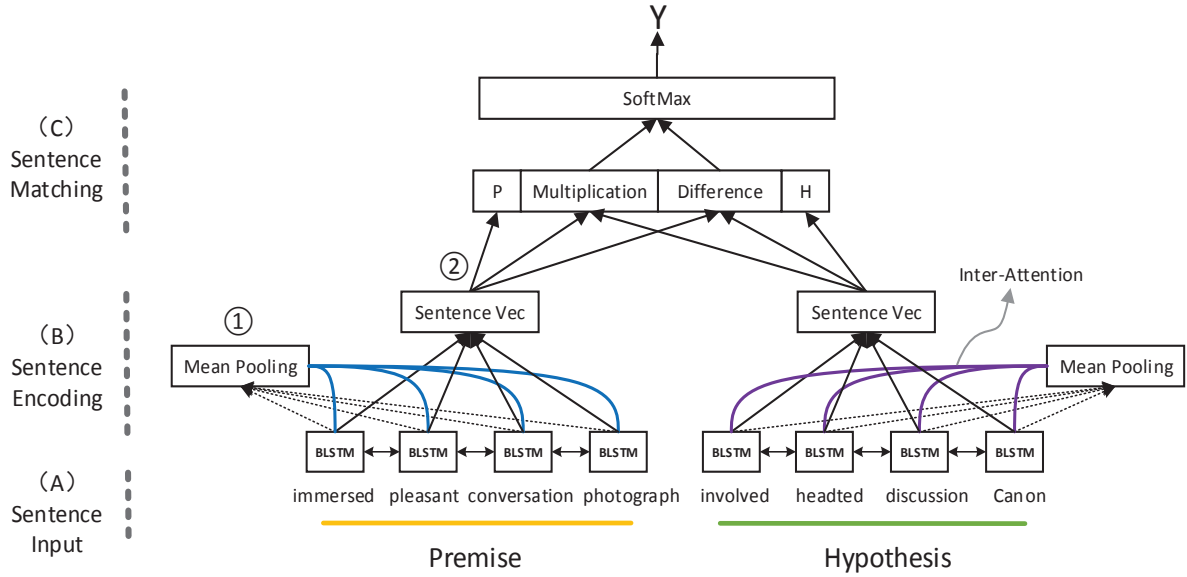}
  \caption{Architecture of Bidirectional LSTM model with Inner-Attention}
  \label{arc}
  \end{figure*}

Traditional methods to RTE has been the dominion of classifiers employing hand engineered features, which heavily relied on natural language processing pipelines and external resources. Formal reasoning methods ~\cite{bos2005recognising} were also explored by many researchers, but not been widely used  because of its complexity and domain limitations.

Recently published Stanford Natural Language Inference (SNLI\footnote{http://nlp.stanford.edu/projects/snli/}) corpus makes it possible to use deep learning methods to solve RTE problems. So far proposed deep learning approaches can be roughly categorized into two groups: sentence encoding-based models and matching encoding-based models. As the name implies,  the encoding of sentence is the core of former methods, while the latter methods directly model the relation between two sentences and didn't generate sentence representations at all.

In view of universality, we focused our efforts on sentence encoding-based model. Existing methods of this kind including: LSTMs-based model, GRUs-based model, TBCNN-based model and SPINN-based model. Single directional LSTMs and GRUs suffer a weakness of not utilizing the contextual information from the future tokens and Convolutional Neural Networks didn't make full use of information contained in word order. Bidirectional LSTM utilizes both the previous and future context by processing the sequence on two directions which helps to address the drawbacks mentioned above. ~\cite{tan2015lstm}

A recent work by ~\cite{rocktaschel2015reasoning} improved the performance by applying a neural attention model that didn't yield sentence embeddings.

In this paper, we proposed a unified deep learning framework for recognizing textual entailment which dose not require any feature engineering, or external resources. The basic model is based on building biLSTM models on both premises and hypothesis. The basic mean pooling encoder can roughly form a intuition about what this sentence is talking about. Obtained this representation, we extended this model by utilize an Inner-Attention mechanism on both sides. This mechanism helps generate more accurate and focused sentence representations for classification. In addition, we introduced a simple effective input strategy that get ride of same words in hypothesis and premise, which further boosts our performance. Without parameter tuning, we improved the art-of-the-state performance of sentence encoding-based model by nearly 2\%.

\section{Our approach}
In our work, we treated RTE task as a supervised three-way classification problem. The overall architecture of our model is shown in Figure~\ref{arc}. The design of this model we follow the idea of Siamese Network, that the two identical sentence encoders share the same set of weights during training, and the two sentence representations then combined together to generated a "relation vector" for classification. As we can see from the figure, the model mainly consists of three parts. From top to bottom were: (A). The sentence input module; (B). The sentence encoding module; (C). The sentence matching module. We will explain the last two parts in detail in the following subsection. And the sentence input module will be introduced in Section~\ref{ssec:InputStrategy}.

\subsection{Sentence Encoding Module}
Sentence encoding module is the fundamental part of this model. To generate better sentence representations, we employed a two-step strategy to encode sentences. Firstly, average pooling layer was built on top of word-level biLSTMs to produce sentence vector. This simple encoder combined with the sentence matching module formed the basic architecture of our model. With much less parameters, this basic model alone can outperformed art-of-state method by a small margin. (refer to Table~\ref{ComPerf}). Secondly, attention mechanism was employed on the same sentence, instead of using target sentence representation to attend words in source sentence, we used the representation generated in previous stage to attend words appeared in the sentence itself, which results in a similar distribution with other attention mechanism weights. More attention was given to important words.\footnote{Recently, ~\cite{yang2016hierarchical} proposed a Hierarchical Attention model on the task of document classification also used for but the target representation in attention their mechanism is randomly initialized.}

The idea of "Inner-attention" was inspired by the observation that when human read one sentence, people usually can roughly form an intuition about which part of the sentence is more important according past experience. And we implemented this idea using attention mechanism in our model. The attention mechanism is formalized as follows:
\begin{eqnarray*}
& M = tanh(W^yY + W^hR_{ave}\otimes e_L) \\
& \alpha = softmax(w^TM) \\
& R_{att} = Y\alpha^T
\end{eqnarray*}
where $Y$ is a matrix consisting of output vectors of biLSTM, $R_{ave}$ is the output of mean pooling layer, $\alpha$ denoted the attention vector and $R_{att}$ is the \emph{attention-weighted} sentence representation.
\subsection{Sentence Matching Module}
Once the sentence vectors are generated. Three matching methods were applied to extract relations between premise and hypothesis.
\begin{itemize}
\setlength{\itemsep}{0pt}
\setlength{\parsep}{0pt}
\setlength{\parskip}{0pt}
\item Concatenation of the two representations
\item Element-wise product
\item Element-wise difference
\end{itemize}
This matching architecture was first used by ~\cite{mou2015recognizing}.
Finally, we used a SoftMax layer over the output of a non-linear projection of the generated matching vector for classification.
\section{Experiments}
\label{sec:Experiments}
\subsection{DataSet}
To evaluate the performance of our model, we conducted our experiments on Stanford Natural Language Inference (SNLI) corpus ~\cite{bos2005recognising}. At 570K pairs, SNLI is two orders of magnitude larger than all other resources of its type. The dataset is constructed by crowdsourced efforts, each sentence written by humans. The target labels comprise three classes: Entailment, Contradiction, and Neutral (two irrelevant sentences). We applied the standard
train/validation/test split, containing 550k, 10k, and 10k samples, respectively.

\subsection{Parameter Setting}
The training objective of our model is cross-entropy loss, and we use minibatch SGD with the Rmsprop ~\cite{tieleman2012lecture} for optimization. The batch size is 128. A dropout layer was applied in the output of the network with the dropout rate set to 0.25. In our model, we used pretrained 300D Glove 840B vectors~\cite{pennington2014glove} to initialize the word embedding. Out-of-vocabulary words in the training set are randomly initialized by sampling values uniformly from (−0.05, 0.05). All of these embedding are not updated during training . We didn't tune representations of words for two reasons:
1. To reduced the number of parameters needed to train.
2. Keep their representation stays close to unseen similar words in inference time, which improved the model's generation ability.
The model is implemented using open-source framework Keras.\footnote{http://keras.io/}

\subsection{The Input Strategy}
\label{ssec:InputStrategy}

In this part, we investigated four strategies to modify the input on our basic model which helps us increase performance, the four strategies are:
\begin{itemize}
\setlength{\itemsep}{0pt}
\setlength{\parsep}{0pt}
\setlength{\parskip}{0pt}
\item Inverting Premises~\cite{sutskever2014sequence}
\item Doubling Premises~\cite{zaremba2014learning}
\item Doubling Hypothesis
\item Differentiating Inputs (Removing same words appeared in premises and hypothesis)
\end{itemize}

Experimental results were illustrated in Table~\ref{InputSgy}. As we can see from it, doubling hypothesis and differentiating inputs both improved our model's performance.While the hypothesises usually much shorter than premises, doubling hypothesis may absorb this difference and emphasize
the meaning twice via this strategy. Differentiating input strategy forces the model to focus on different part of the two sentences
which may help the classification for Neutral and Contradiction examples as we observed that our model tended to assign unconfident instances to
Entailment. And the original input sentences appeared in Figure~\ref{arc} are:
\begin{small}

\textbf{Premise:} \emph{Two man in polo shirts  and tan pants immersed in a pleasant conversation about photograph.}

\textbf{Hypothesis:} \emph{Two man in polo shirts and tan pants involved in a heated discussion about Canon.}

\textbf{Label:} \emph{Contradiction}

\end{small}
While most of the words in this pair of sentences are same or close in semantic, It is hard for model to distinguish the difference between them, which resulted in labeling it with Neutral or Entailment. Through differentiating inputs strategy, this kind of problems can be solved.
\begin{table}[]
\centering
\begin{tabular}{|l|c|}
\hline
\multicolumn{1}{|c|}{\textbf{Input Strategy}} & \textbf{Test Acc.} \\ \hline
Original Sequences                            & 83.24\%            \\ \hline
Inverting Premises                            & 82.60\%            \\ \hline
Doubling Premises                             & 83.66\%            \\ \hline
Doubling Hypothesis                           & 82.83\%            \\ \hline
Differentiating Inputs                        & \textbf{83.72\%}   \\ \hline
\end{tabular}
\caption{Comparison of different input strategies}
\label{InputSgy}
\end{table}
\subsection{Comparison Methods}
In this part, we compared our model against the following art-of-the-state baseline approaches:
\begin{small}
\begin{itemize}
\setlength{\itemsep}{0pt}
\setlength{\parsep}{0pt}
\setlength{\parskip}{0pt}
  \item \textbf{LSTM enc:} 100D LSTM encoders + MLP. ~\cite{bowman2015large}
  \item \textbf{GRU enc:} 1024D GRU encoders + skip-thoughts + cat, -. ~\cite{vendrov2015order}
  \item \textbf{TBCNN enc:} 300D Tree-based CNN encoders + cat, $\circ$ , -. ~\cite{mou2015recognizing}
  \item \textbf{SPINN enc:} 300D SPINN-NP encoders + cat, $\circ$ , -. ~\cite{bowman2016fast}
  \item \textbf{Static-Attention:} 100D LSTM + static attention. ~\cite{rocktaschel2015reasoning}
  \item \textbf{WbW-Attention:} 100D LSTM + word-by-word attention. ~\cite{rocktaschel2015reasoning}
\end{itemize}
\end{small}
The \emph{\textbf{cat}} refers to concatenation, \emph{\textbf{-}} and \textbf{$\circ$} denote element-wise difference and product, respectively.
Much simpler and easy to understand.
\begin{table}[]
\centering
\begin{tabular}{|l|c|c|}
\hline
\textbf{Model}             & \multicolumn{1}{l|}{\textbf{Params}} & \multicolumn{1}{l|}{\textbf{Test Acc.}} \\ \hline
\multicolumn{3}{|l|}{\textit{\textbf{Sentence encoding-based models}}}                                      \\ \hline
LSTM enc                   & 3.0M                                 & 80.6\%                                  \\ \hline
GRU enc                    & 15M                                  & 81.4\%                                  \\ \hline
TBCNN enc                  & 3.5M                                 & 82.1\%                                  \\ \hline
SPINN enc                  & 3.7M                                 & 83.2\%                                  \\ \hline
Basic model                & \textbf{2.0M}                        & 83.3\%                                  \\
\textit{+ Inner-Attention} & 2.8M                                 & 84.2\%                                  \\
\textit{+ Diversing Input} & 2.8M                                 & \textbf{85.0\%}                         \\ \hline
\multicolumn{3}{|l|}{\textit{\textbf{Other neural network models}}}                                         \\ \hline
Static-Attention           & 242K                                 & 82.4\%                                  \\ \hline
WbW-Attention              & 252K                                 & 83.5\%                                  \\ \hline
\end{tabular}
\caption{Performance comparison of different models on SNLI.}
\label{ComPerf}
\end{table}

\subsection{Results and Qualitative Analysis}
\begin{figure}[htbp]
  \centering
  \includegraphics[height=4cm, width=7.6cm]{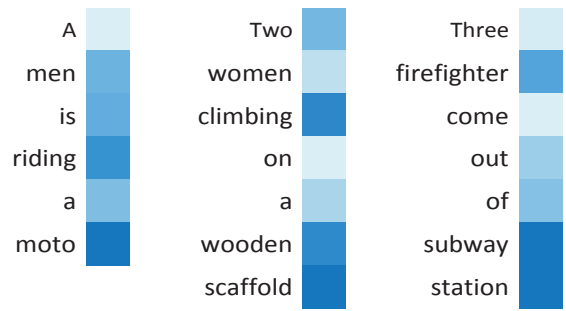}
  \caption{Inner-Attention Visualizations.}
  \label{VisAtt}
\end{figure}

Although the classification of RTE example is not solely relying on representations obtained from attention, it is still instructive to analysis Inner-Attention mechanism as we witnessed a large performance increase after employing it. We hand-picked several examples from the dataset to visualize. In order to make the weights more discriminated, we didn't use a uniform colour atla cross sentences. That is, each sentence have its own color atla, the lightest color and the darkest color denoted the smallest attention weight the biggest value within the sentence, respectively. Visualizations of Inner-Attention on these examples are depicted in Figure~\ref{VisAtt}.

We observed that more attention was given to Nones, Verbs and Adjectives. This conform to our experience that these words are more semantic richer than function words. While mean pooling regarded each word of equal importance, the attention mechanism helps re-weight words according to their importance. And more focused and accurate sentence representations were generated based on produced attention vectors.

\section{Conclusion and Future work}
In this paper, we proposed a bidirectional LSTM-based model with Inner-Attention to solve the RTE problem. We come up with an idea to utilize attention mechanism within sentence which can teach itself to attend words without the information from another one. The Inner-Attention mechanism helps produce more accurate sentence representations through attention vectors. In addition, the simple effective diversing input strategy introduced by us further boosts our results. And this model can be easily adapted to other sentence-matching models.
Our future work including:
\begin{enumerate}
\setlength{\itemsep}{0pt}
\setlength{\parsep}{0pt}
\setlength{\parskip}{0pt}
\item Employ this architecture on other sentence-matching tasks such as Question Answer, Paraphrase and Sentence Text Similarity etc.
\item Try more heuristics matching methods to make full use of the sentence vectors.
\end{enumerate}
\section*{Acknowledgments}
We thank all anonymous reviewers for their hard work!

\bibliography{emnlp2016}

\begin{thebibliography}{}

\bibitem[\protect\citename{Bos and Markert}2005]{bos2005recognising}
Johan Bos and Katja Markert.
\newblock 2005.
\newblock Recognising textual entailment with logical inference.
\newblock In {\em Proceedings of the conference on Human Language Technology
  and Empirical Methods in Natural Language Processing}, pages 628--635.
  Association for Computational Linguistics.

\bibitem[\protect\citename{Bowman \bgroup et al.\egroup }2015]{bowman2015large}
Samuel~R Bowman, Gabor Angeli, Christopher Potts, and Christopher~D Manning.
\newblock 2015.
\newblock A large annotated corpus for learning natural language inference.
\newblock {\em arXiv preprint arXiv:1508.05326}.

\bibitem[\protect\citename{Bowman \bgroup et al.\egroup }2016]{bowman2016fast}
Samuel~R Bowman, Jon Gauthier, Abhinav Rastogi, Raghav Gupta, Christopher~D
  Manning, and Christopher Potts.
\newblock 2016.
\newblock A fast unified model for parsing and sentence understanding.
\newblock {\em arXiv preprint arXiv:1603.06021}.

\bibitem[\protect\citename{Mou \bgroup et al.\egroup }2015]{mou2015recognizing}
Lili Mou, Men Rui, Ge~Li, Yan Xu, Lu~Zhang, Rui Yan, and Zhi Jin.
\newblock 2015.
\newblock Recognizing entailment and contradiction by tree-based convolution.
\newblock {\em arXiv preprint arXiv:1512.08422}.

\bibitem[\protect\citename{Pennington \bgroup et al.\egroup
  }2014]{pennington2014glove}
Jeffrey Pennington, Richard Socher, and Christopher~D Manning.
\newblock 2014.
\newblock Glove: Global vectors for word representation.
\newblock In {\em EMNLP}, volume~14, pages 1532--1543.

\bibitem[\protect\citename{Rockt{\"a}schel \bgroup et al.\egroup
  }2015]{rocktaschel2015reasoning}
Tim Rockt{\"a}schel, Edward Grefenstette, Karl~Moritz Hermann, Tom{\'a}{\v{s}}
  Ko{\v{c}}isk{\`y}, and Phil Blunsom.
\newblock 2015.
\newblock Reasoning about entailment with neural attention.
\newblock {\em arXiv preprint arXiv:1509.06664}.

\bibitem[\protect\citename{Sutskever \bgroup et al.\egroup
  }2014]{sutskever2014sequence}
Ilya Sutskever, Oriol Vinyals, and Quoc~V Le.
\newblock 2014.
\newblock Sequence to sequence learning with neural networks.
\newblock In {\em Advances in neural information processing systems}, pages
  3104--3112.

\bibitem[\protect\citename{Tan \bgroup et al.\egroup }2015]{tan2015lstm}
Ming Tan, Bing Xiang, and Bowen Zhou.
\newblock 2015.
\newblock Lstm-based deep learning models for non-factoid answer selection.
\newblock {\em arXiv preprint arXiv:1511.04108}.

\bibitem[\protect\citename{Tieleman and Hinton}2012]{tieleman2012lecture}
Tijmen Tieleman and Geoffrey Hinton.
\newblock 2012.
\newblock Lecture 6.5-rmsprop.
\newblock {\em COURSERA: Neural networks for machine learning}.

\bibitem[\protect\citename{Vendrov \bgroup et al.\egroup
  }2015]{vendrov2015order}
Ivan Vendrov, Ryan Kiros, Sanja Fidler, and Raquel Urtasun.
\newblock 2015.
\newblock Order-embeddings of images and language.
\newblock {\em arXiv preprint arXiv:1511.06361}.

\bibitem[\protect\citename{Yang \bgroup et al.\egroup
  }2016]{yang2016hierarchical}
Zichao Yang, Diyi Yang, Chris Dyer, Xiaodong He, Alex Smola, and Eduard Hovy.
\newblock 2016.
\newblock Hierarchical attention networks for document classification.
\newblock In {\em Proceedings of the 2016 Conference of the North American
  Chapter of the Association for Computational Linguistics: Human Language
  Technologies}.

\bibitem[\protect\citename{Zaremba and Sutskever}2014]{zaremba2014learning}
Wojciech Zaremba and Ilya Sutskever.
\newblock 2014.
\newblock Learning to execute.
\newblock {\em arXiv preprint arXiv:1410.4615}.

\end{thebibliography}
\bibliographystyle{emnlp2016}

\end{document}